\documentclass[conference]{IEEEtran}
\IEEEoverridecommandlockouts
\usepackage{amsmath,amssymb,amsfonts}
\usepackage{algorithmic}
\usepackage{graphicx}
\usepackage{textcomp}
\usepackage{xcolor}
\usepackage{booktabs}
\usepackage[backend=bibtex]{biblatex}
\begin{document}

\title{Analysis of Bias in Deep Learning Facial Beauty Regressors}

\author{\IEEEauthorblockN{1\textsuperscript{st} Chandon Hamel}
\IEEEauthorblockA{\textit{Anderson College of Business and Computing} \\
\textit{Regis University}\\
Denver, Colorado \\
chamel@regis.edu}
~\\
\and
\IEEEauthorblockN{2\textsuperscript{nd} Mike Busch}
\IEEEauthorblockA{\textit{Anderson College of Business and Computing} \\
\textit{Regis University}\\
Denver, Colorado \\
mbusch@regis.edu}
}

\maketitle

\begin{abstract}
Bias can be introduced to AI systems even from seemingly balanced sources, and AI facial beauty prediction is subject to ethnicity-based bias. This work sounds warnings about AI's role in shaping aesthetic norms while providing potential pathways toward equitable beauty technologies through comparative analysis of models trained on SCUT-FBP5500 and MEBeauty datasets. Employing rigorous statistical validation (Kruskal-Wallis H-tests, post hoc Dunn analyses). It is demonstrated that both models exhibit significant prediction disparities across ethnic groups $(p < 0.001)$, even when evaluated on the balanced FairFace dataset. Cross-dataset validation shows algorithmic amplification of societal beauty biases rather than mitigation based on prediction and error parity. The findings underscore the inadequacy of current AI beauty prediction approaches, with only 4.8-9.5\% of inter-group comparisons satisfying distributional parity criteria. Mitigation strategies are proposed and discussed in detail. 
\end{abstract}

\begin{IEEEkeywords}
Facial Beauty Prediction, Algorithmic Bias, Fairness in AI, Computer Vision, Deep Learning, Statistical Analysis
\end{IEEEkeywords}

\section{Introduction}
Beauty is in the eye of the beholder, unless the beholder has no eyes. As computer vision systems become ubiquitous across industries, their growing influence on human self-perception warrants scrutiny, particularly in beauty assessment applications. 

The deployment of facial beauty prediction models potentially impacts critical domains including:
\begin{itemize}
    \item Automated talent screening in acting/modeling agencies
    \item Beauty-score-targeted marketing by cosmetic companies  
    \item Post-surgical outcome simulation in aesthetic clinics
    \item AI-curated profile picture selection on social platforms
\end{itemize}

These applications risk codifying and amplifying subjective beauty standards through algorithmic reproduction. When such systems exhibit ethnic bias, they perpetuate harmful stereotypes and create unequal access to opportunities tied to perceived attractiveness. 

This study conducts a comparative analysis of ethnic bias in facial beauty assessment models trained on two publicly available datasets: SCUT-FBP5500~\cite{liang2017SCUT} and MEBeauty~\cite{lebedeva2021mebeauty}. Its purpose is to analyze disparities in prediction and error distributions across ethnic groups, thereby assessing the extent and nature of bias in models trained to predict beauty scores.

\section{Motivation}

\subsection{Related Works}

Following the bias analysis framework demonstrated in hiring algorithms by \textcite{Quer2024Analyzing}, this study adapts their evaluation approach to continuous beauty score prediction through two key fairness metrics:

\begin{enumerate}
    \item Distributional Parity
    \begin{itemize}
        \item Analogous to Statistical Parity for classification
        \item Assessed via Mann-Whitney U/Kruskal-Wallis tests on prediction distributions
        \item Reveals systemic score disparities across demographic groups
    \end{itemize}
    \item Error Parity
    \begin{itemize}
        \item Analogous to Equal Opportunity Difference
        \item Evaluated through non-parametric tests on error distributions
        \item Identifies differential model performance between groups
    \end{itemize}
\end{enumerate}

These metrics build on a bias mitigation framework from \textcite{Feldman2021EndToEnd}, extending their fairness validation techniques from discrete classification to continuous regression tasks. Where \textcite{Quer2024Analyzing} focused on categorical outcomes in talent matching, this work addresses fairness in subjective beauty score prediction.

The importance of this analysis is underscored by mounting evidence of representational harms in generative AI systems. Recent studies reveal that image-generation models:

\begin{itemize}
    \item Produce less diverse outputs than human-curated content~\cite{Bogdanova2024Diversity}
    \item Underrepresent women and People of Color in depictions of power and success \cite{Gengler2024Sexism}
    \item Systematically whitify non-white faces in image-to-image transformations \cite{Yang2025Racial}
\end{itemize}

These findings suggest beauty prediction models risk amplifying similar biases through automated scoring. This study bridges this critical gap by examining how regression-based scoring systems may perpetuate harmful norms through biased prediction patterns.

\subsection{Ethical Implications}

The ethical concerns regarding AI-driven beauty standards are evident in both empirical research and philosophical analysis. According to a recent survey by \textcite{Ilyas2024AIBeauty}, 50\% of respondents reported that exposure to AI-generated beauty standards negatively impacted their self-esteem, while 70\% agreed that such standards promote unrealistic cultural and social ideals. Furthermore, 82\% of respondents felt that AI-based beauty images are less inclusive in promoting diversity across cultures. These findings underscore growing concerns about the effects of AI on self-worth and cultural inclusivity.

Philosophical perspectives further illuminate the risks posed by algorithmic beauty standards. As \textcite{Zhou2024AIBeauty} argues, the introduction of AI into the domain of beauty risks "warping natural beauty standards and making the original human appear increasingly imperfect." Zhou situates this concern within a broader historical context, noting that while philosophers like Plato conceived of beauty as an abstract, eternal ideal, and Hume emphasized its subjectivity, AI systems operationalize beauty in ways that are both highly specific and potentially exclusionary. By learning from biased or narrow data, AI models may reinforce harmful stereotypes, objectify marginalized groups, and amplify unattainable ideals, especially for women and people of color.

In summary, the ethical implications of AI in beauty prediction extend beyond technical fairness to encompass broader social harms. These include the perpetuation of unrealistic and exclusionary beauty ideals, the erosion of self-esteem, and the marginalization of diverse cultural expressions of beauty. Addressing these challenges requires not only technical interventions to improve model fairness, but also industry-wide commitments to transparency, inclusivity, and the responsible use of AI in shaping societal standards.

\section{Experiment}

This study employs a three-phase computational pipeline to assess ethnic bias in facial beauty prediction models, structured as follows:

\begin{enumerate}
    \item \textbf{Model Development:} Fine-tune ResNet-152 architectures \parencite{He2016Deep} on the SCUT-FBP5500~\parencite{liang2017SCUT} and MEBeauty~\parencite{lebedeva2021mebeauty} datasets separately, maintaining dataset isolation to preserve their inherent bias profiles.
    
    \item \textbf{Model Evaluation:} Generate beauty score predictions using each trained model on the alternate labeled dataset not used in training and the FairFace~\parencite{karkkainenfairface} dataset---a large, demographically diverse, but unlabeled set of face images. This approach enables assessment of model generalization and bias across both curated and real-world data distributions.
    
    \item \textbf{Bias Quantification:} Apply non-parametric statistical tests (Mann-Whitney U and Kruskal-Wallis) to prediction distributions and errors across ethnic groups.
\end{enumerate}

This cross-dataset validation approach enables detection of both dataset-specific biases and model-generalization fairness issues. Subsequent sections detail implementation specifics for reproducibility.

\subsection{Datasets}

\subsubsection{SCUT-FBP5500~\parencite{liang2017SCUT}}

The SCUT-FBP5500 dataset contains 5,500 high-quality frontal face images, divided into four demographic groups: 2,000 Asian females, 2,000 Asian males, 750 Caucasian females, and 750 Caucasian males. Most images were sourced from the internet and each was assigned a beauty score in the range [1, 5], computed as the average rating from 60 volunteers. 


\subsubsection{MEBeauty~\parencite{lebedeva2021mebeauty}}

The MEBeauty dataset comprises 2,550 images representing Black, Asian, Caucasian, Hispanic, Indian, and Mideastern female and male faces. Each image is rated on a [1, 10] beauty scale by approximately 300 individuals from diverse cultural and social backgrounds.


\subsubsection{FairFace Training Subset~\parencite{karkkainenfairface}}

The FairFace training subset consists of 86,744 images spanning seven racial groups: White, Black, Indian, East Asian, Southeast Asian, Middle Eastern, and Latino. Designed to mitigate race bias, FairFace emphasizes balanced representation across these groups. Images were collected from the YFCC-100M Flickr dataset~\parencite{Thomee_2016} and are annotated with race, gender, and age group labels. 

\subsection{Preprocessing}

\subsubsection{Metadata}

For each dataset, a tabular metadata file was constructed with columns for the image filename, the subject's race, and, for the SCUT-FBP5500 and MEBeauty datasets, the corresponding labeled beauty score. To ensure consistency across datasets, all beauty scores were normalized to the range [0, 1] prior to model training.

\subsubsection{Images}

\begin{figure}[t]
    \centering
    \includegraphics[width=\linewidth]{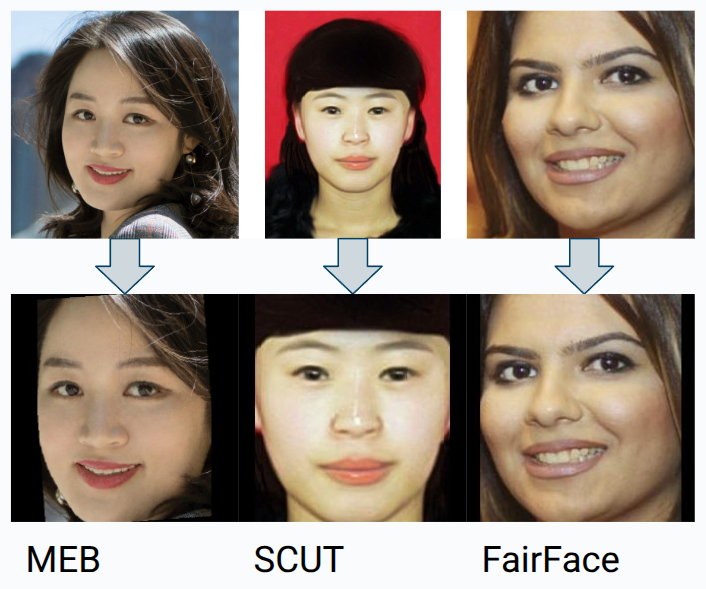}
    \caption{Examples of face cropping and padding using MTCNN.}
    \label{fig:crop}
\end{figure}

Facial regions were extracted from each image using the Multi-task Cascaded Convolutional Networks (MTCNN) face detector \parencite{Zhang_2016}. This process removed background artifacts and ensured that only the subject's face was retained. To maintain aspect ratio and avoid distortion, images were padded with black borders to produce square images. MTCNN failed to detect faces in 2,015 images (2.3\%) from the FairFace dataset, which were subsequently excluded from further analysis. Fig.~\ref{fig:crop} illustrates examples of the face cropping and padding transformation.

\subsection{Model Training}

\subsubsection{Data Loading}

Training pipelines were implemented using PyTorch and PyTorch Lightning. A custom \texttt{Dataset} class performed the following operations for training data:

\begin{itemize}
    \item \textbf{Augmentation:} Random horizontal flips, $\pm10^\circ$ rotations, and resized crops (80-100\% scale) of training images
    \item \textbf{Normalization:} Pixel values scaled using ImageNet means ($\mu = [0.485, 0.456, 0.406]$) and standard deviations ($\sigma = [0.229, 0.224, 0.225]$) \parencite{5206848}
\end{itemize}

Datasets were split into training ($\frac{2}{3}$), validation ($\frac{2}{9}$), and test ($\frac{1}{9}$) subsets. Separate dataloaders were created for training/validation/test splits of the source dataset and external evaluation datasets (cross-dataset and FairFace). Batch sizes were optimized per dataset---64 for SCUT-FBP5500 and 32 for MEBeauty---for computational efficiency and stability during training.

\subsubsection{ResNet-152 Fine-tuning}

The base ResNet-152 architecture \parencite{He2016Deep} was adapted for regression by replacing its final fully connected layer with a single-output linear layer. Training proceeded in three progressive phases:

\begin{enumerate}
    \item \textbf{Frozen Backbone:} Only the final layer trained using Adam optimizer ($\eta = 3 \times 10^{-3}$), ReduceLROnPlateau scheduler (factor=0.5, patience=2 epochs), and early stopping (patience=5 epochs) 
    
    \item \textbf{Unfreeze Conv5:} Resumed training with conv5 block unfrozen
    
    \item \textbf{Unfreeze Conv4:} Final training phase with conv4 block unfrozen
\end{enumerate}

Models were continuously monitored via validation MSE loss, with restoration to best-performing checkpoints between phases. 

\subsection{Bias Analysis}

\subsubsection{Model Performance}

Table~\ref{table:performance} presents the performance metrics for both models, evaluated on their respective test sets and cross-validated on the alternative dataset. The SCUT-trained model achieved a test Mean Squared Error (MSE) of 0.008, while the MEBeauty-trained model reached 0.013. These results are competitive with state-of-the-art performance on facial beauty prediction tasks.

\begin{table}[b!]
    \caption{Model performance measured by Mean Squared Error (MSE) on normalized [0,1] scale. Test MSE represents performance on the held-out test set from the training dataset. Cross-dataset MSE shows performance when SCUT-trained model is evaluated on MEBeauty and vice versa.}
    \centering
    \begin{tabular}{|l|c|c|} 
    \hline 
    \textbf{Model} & \textbf{Test MSE} & \textbf{Cross-dataset MSE} \\
    \hline
    SCUT-trained & 0.008 & 0.024 \\
    MEBeauty-trained & 0.013 & 0.028 \\
    \hline
    \end{tabular}
    \label{table:performance}
\end{table}

For contextual comparison, \textcite{liang2017SCUT} reported an optimal RMSE of 0.302 on a [1, 5] scale using ResNeXt-50 \parencite{xie2017aggregatedresidualtransformationsdeep}. This converts to approximately 0.076 on the normalized [0, 1] scale. This study's SCUT-trained model's RMSE of 0.089 demonstrates comparable performance despite using a different, slightly earlier architecture.

Cross-dataset evaluation shows expected performance degradation, with MSE increasing by factors of 3.0 and 2.2 for the SCUT and MEBeauty models respectively. This degradation likely reflects underlying differences in beauty annotation protocols, demographic composition, and visual characteristics between datasets.

\subsubsection{MEBeauty Data Analysis}

\begin{figure}[b]
    \centering
    \includegraphics[width=\linewidth]{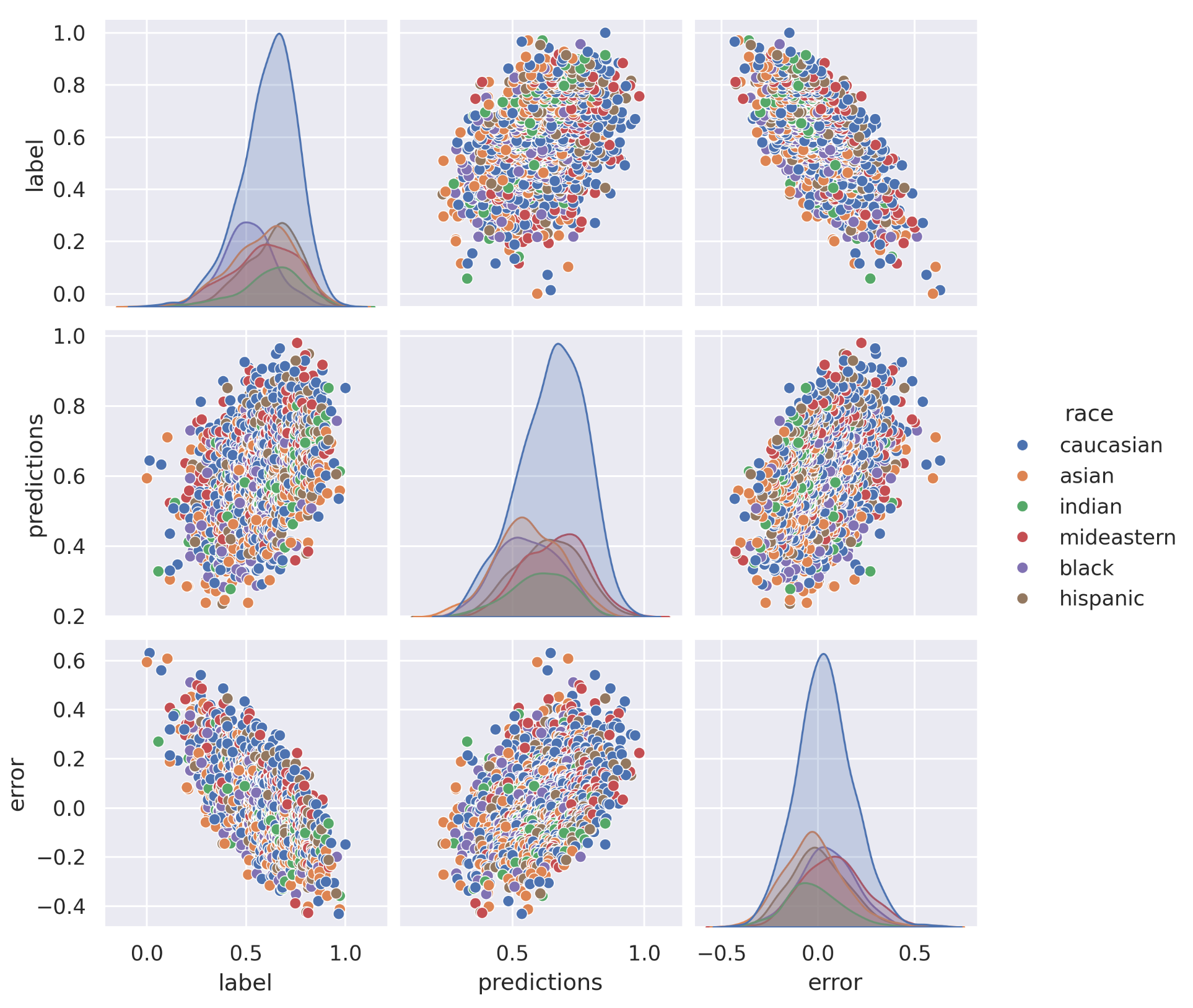}
    \caption{Pairplot of ground truth labels, model predictions, and prediction errors from the SCUT-trained model applied to the MEBeauty dataset, segmented by ethnic group.}
    \label{fig:meb-pairplot}
\end{figure}

Fig.~\ref{fig:meb-pairplot} presents a pairplot visualization of the MEBeauty dataset, showing distributions of ground truth labels, model predictions from the SCUT-trained model, and prediction errors across ethnic groups. Visual inspection reveals notable distributional differences that suggest potential bias in both the dataset annotations and model predictions.

Quantitative analysis of the SCUT-trained model's performance on MEBeauty data is summarized in Table~\ref{table:meb-data}. The results reveal substantial disparities in beauty score predictions across ethnic groups. Caucasian and Middle Eastern faces received significantly higher mean predicted beauty scores (0.65 and 0.66 respectively) compared to Asian and Black faces (both 0.56). Error analysis further indicates systematic bias, with Middle Eastern faces showing the largest positive error (0.07), suggesting consistent overestimation of beauty scores for this group.

Statistical validation via Kruskal-Wallis tests strongly rejects the null hypothesis of equal distribution across ethnic groups for both predictions ($H = 229$, $p = 1.9 \times 10^{-47}$) and errors ($H = 101$, $p = 3.6 \times 10^{-20}$). These findings were further confirmed through permutation-based ANOVA tests, which yielded consistent significance levels ($p = 2 \times 10^{-4}$) for both metrics.

\begin{table}[t]
\centering
\caption{Statistical analysis of predictions and errors from the SCUT-trained model on the MEBeauty dataset, segmented by ethnic group. Lower rows show Kruskal-Wallis test results assessing equality of distributions across groups.}
\label{table:meb-data}
    \begin{tabular}{lcccc}
        \toprule
         & \multicolumn{2}{c}{Predictions} & \multicolumn{2}{c}{Errors} \\
        \cmidrule(lr){2-3} \cmidrule(lr){4-5}
        Race & Mean & Median & Mean & Median \\
        \midrule
        Asian & 0.56 & 0.56 & -0.02 & -0.03 \\
        Black & 0.56 & 0.56 & 0.04 & 0.04 \\
        Caucasian & 0.65 & 0.66 & 0.03 & 0.03 \\
        Hispanic & 0.63 & 0.64 & 0.00 & -0.00 \\
        Indian & 0.60 & 0.61 & -0.02 & -0.03 \\
        Middle Eastern & 0.66 & 0.66 & 0.07 & 0.07 \\
        \midrule
        KW Test Statistic & \multicolumn{2}{c}{229} & \multicolumn{2}{c}{101} \\
        KW Test P-value & \multicolumn{2}{c}{$<10^{-3}$} & \multicolumn{2}{c}{$<10^{-3}$} \\
        \bottomrule
    \end{tabular}
\end{table}

\begin{figure}[b]
    \centering
    \includegraphics[width=\linewidth]{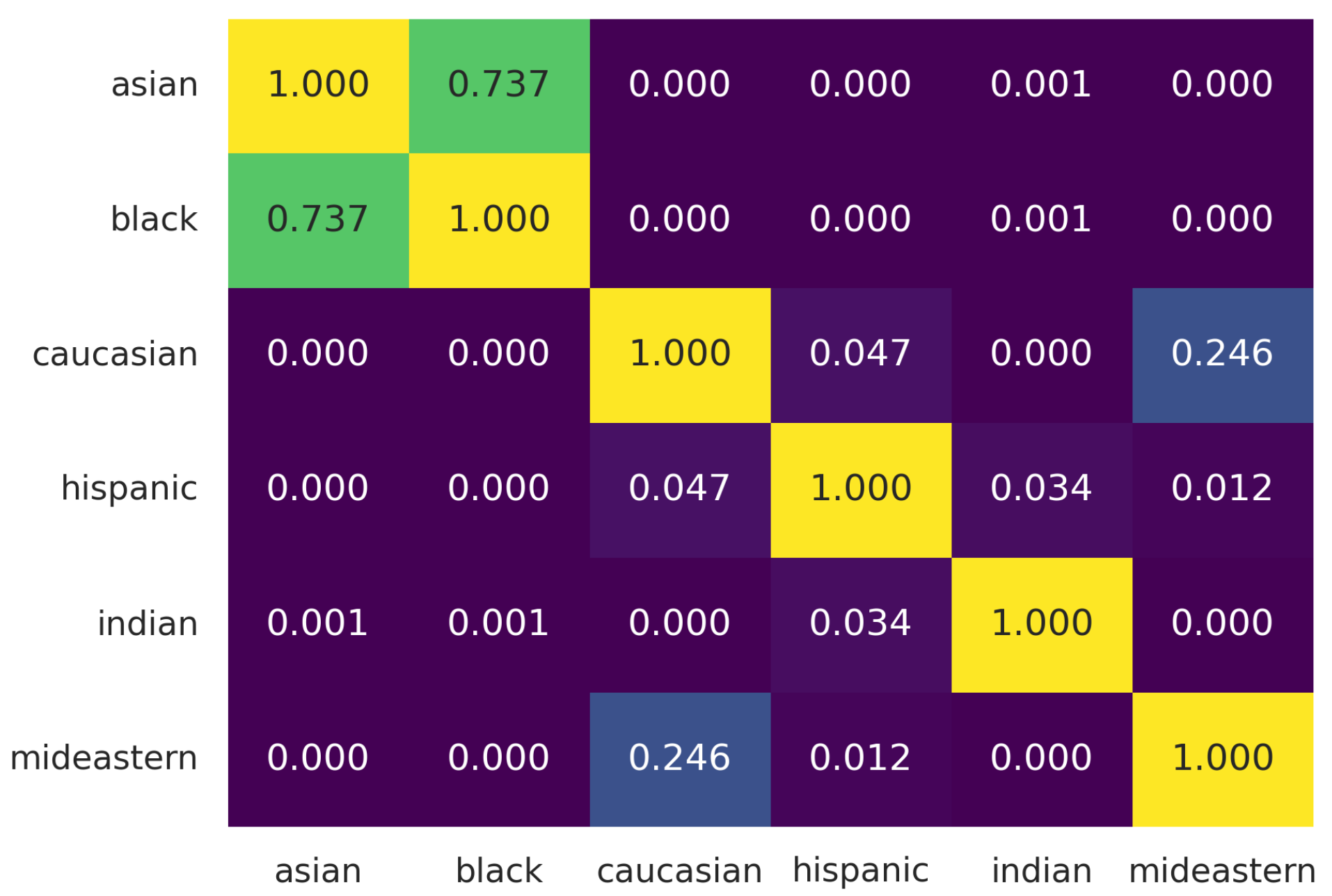}
    \caption{Heatmap of pairwise post hoc Dunn test results for model predictions across ethnic groups (SCUT-trained model on MEBeauty data). Each cell displays the adjusted p-value for the comparison between two groups after Benjamini-Hochberg correction.}
    \label{fig:meb-phd-pred}
\end{figure}

\begin{figure}[tb]
    \centering
    \includegraphics[width=\linewidth]{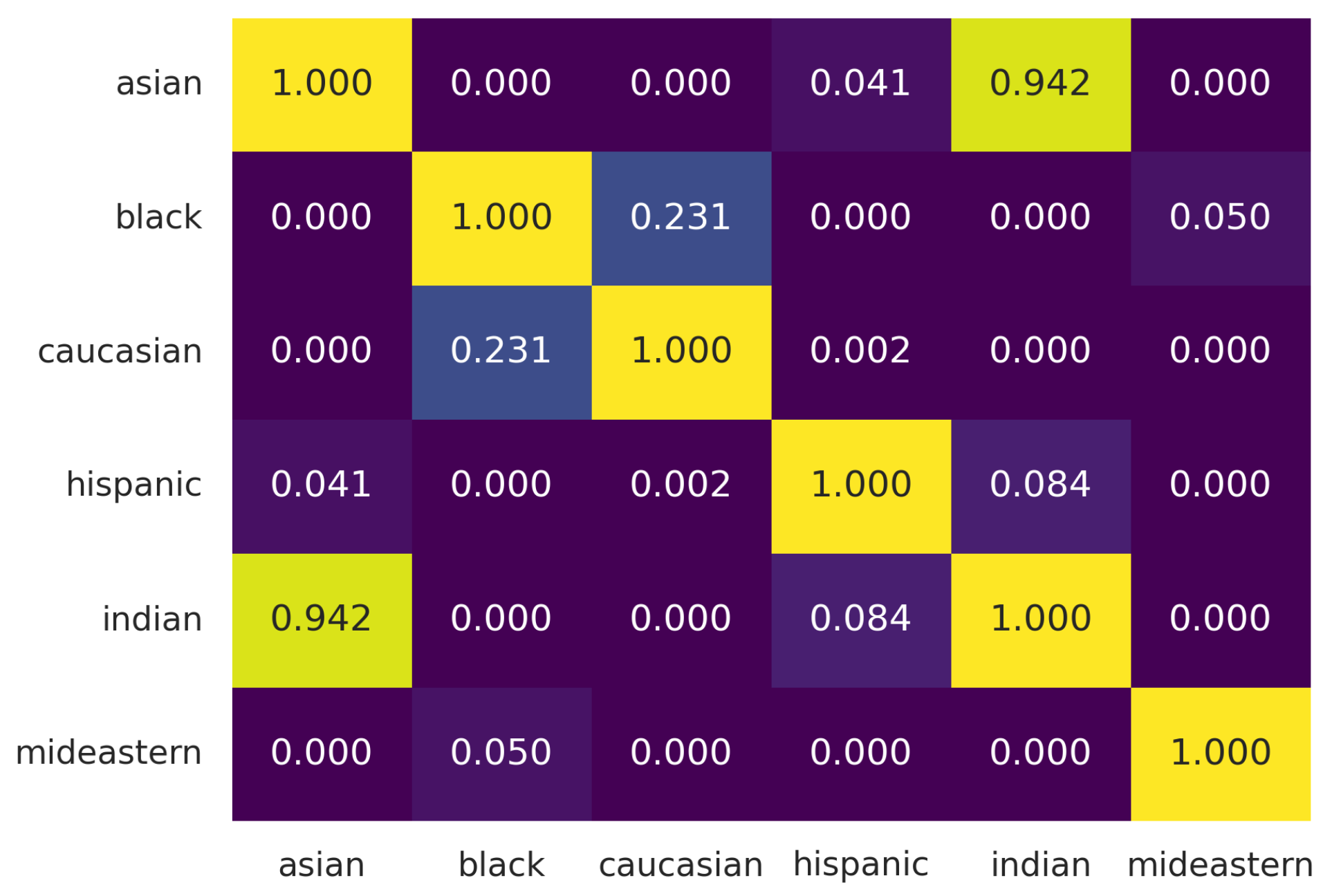}
    \caption{Heatmap of pairwise post hoc Dunn test results for prediction errors across ethnic groups (SCUT-trained model on MEBeauty data). Each cell displays the adjusted p-value for the comparison between two groups after Benjamini-Hochberg correction.}
    \label{fig:meb-phd-error}
\end{figure}

To identify specific inter-group disparities, Post Hoc Dunn tests were conducted with Benjamini-Hochberg p-value adjustment to control for multiple comparisons. Fig.~\ref{fig:meb-phd-pred} visualizes these pairwise comparisons for predictions, revealing that only 2 out of 15 ethnic group pairs (13.3\%) satisfy the Distributional Parity criterion at $\alpha = 0.05$. Similarly, Fig.~\ref{fig:meb-phd-error} shows that only 3 pairs (20\%) meet the Error Parity standard. These results provide strong statistical evidence of ethnic bias in the SCUT-trained model when applied to the MEBeauty dataset.

\subsubsection{SCUT Data Analysis}

\begin{figure}[tb]
    \centering
    \includegraphics[width=\linewidth]{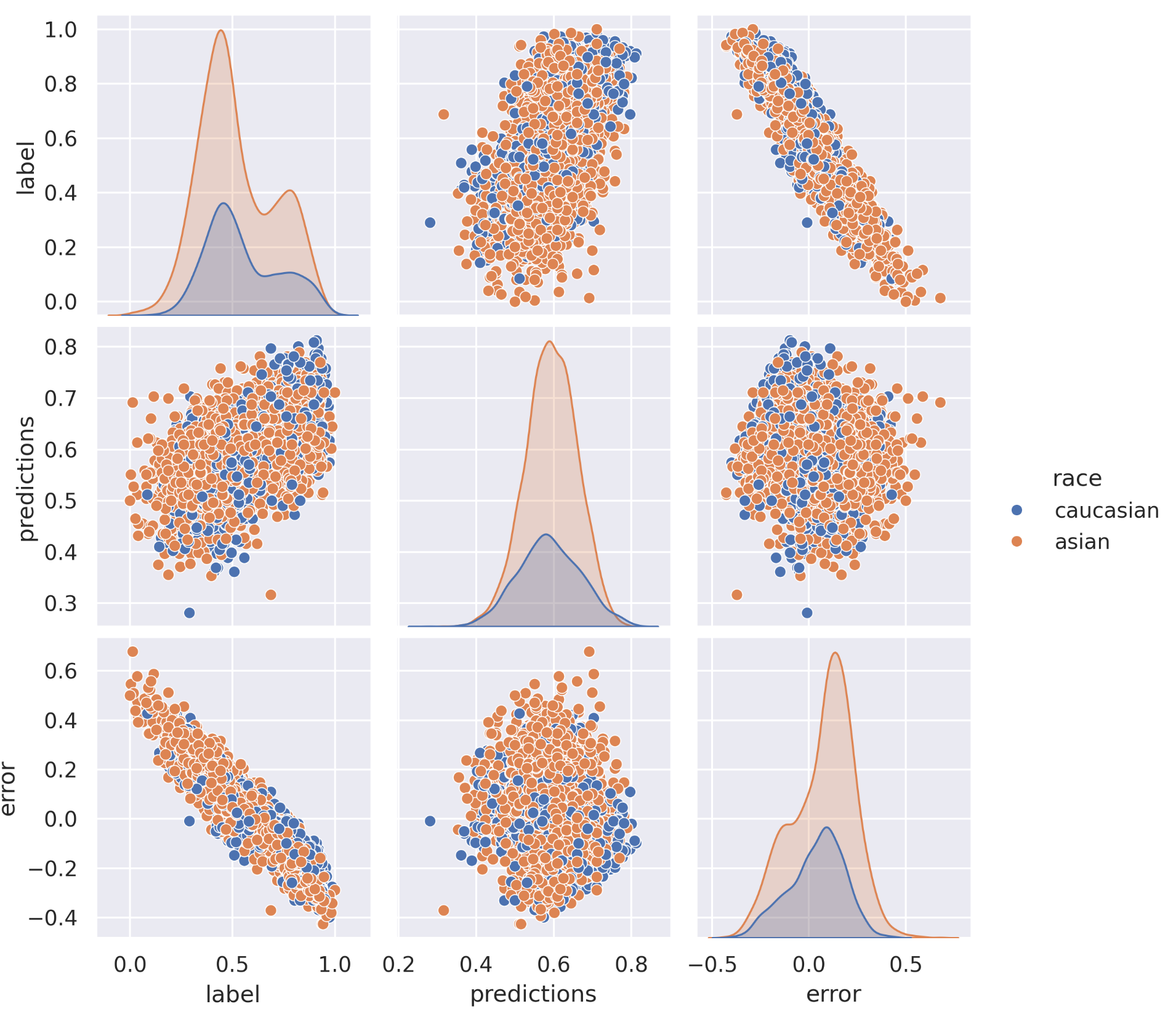}
    \caption{Pairplot of ground truth labels, model predictions, and prediction errors from the MEBeauty-trained model applied to the SCUT-FBP5500 dataset, segmented by ethnic group (Asian and Caucasian).}
    \label{fig:scut-pairplot}
\end{figure}

Fig.~\ref{fig:scut-pairplot} presents a pairplot visualization of the SCUT-FBP5500 dataset, illustrating distributions of ground truth labels, model predictions, and prediction errors across Asian and Caucasian ethnic groups. Visual inspection reveals distinct distributional patterns that suggest potential bias in both the dataset annotations and the MEBeauty-trained model's predictions.

Table~\ref{table:scut-data} provides a quantitative analysis of the MEBeauty-trained model's performance on SCUT-FBP5500 data. The results indicate a notable difference in prediction patterns between Asian and Caucasian faces. While mean prediction values are similar (0.5929 for Asian vs. 0.5883 for Caucasian), the error distributions show more substantial disparities. Asian faces exhibit significantly higher mean errors (0.0722) compared to Caucasian faces (0.0380), suggesting systematic overestimation of beauty scores for Asian subjects relative to ground truth labels.

Statistical validation was performed using both Mann-Whitney U (MWU) and Kolmogorov-Smirnov (KS) tests. The MWU test, which assesses differences in distribution medians, rejects the null hypothesis of equal prediction distributions with moderate significance ($U = 2,869,019$, $p = 0.0124$) and strongly rejects equal error distributions ($U = 2,562,921.5$, $p = 7.8 \times 10^{-17}$). The KS test, which is sensitive to differences in distribution shape and spread, provides even stronger evidence of distributional differences for both predictions ($D = 0.0702$, $p = 4.1 \times 10^{-5}$) and errors ($D = 0.146$, $p = 9.8 \times 10^{-21}$).

\begin{table}[t!]
\centering
\caption{Statistical analysis of predictions and errors from the MEBeauty-trained model on the SCUT-FBP5500 dataset, segmented by ethnic group. Lower rows show Mann-Whitney U and Kolmogorov-Smirnov test results assessing equality of distributions across groups.}
\label{table:scut-data}
    \begin{tabular}{lcccc}
        \toprule
         & \multicolumn{2}{c}{Predictions} & \multicolumn{2}{c}{Errors} \\
        \cmidrule(lr){2-3} \cmidrule(lr){4-5}
        Ethnic Group & Mean & Median & Mean & Median \\
        \midrule
        Asian & 0.5929 & 0.5938 & 0.0722 & 0.1004 \\
        Caucasian & 0.5883 & 0.5859 & 0.0380 & 0.0582 \\
        \midrule
        MWU Statistic & \multicolumn{2}{c}{2,869,019} & \multicolumn{2}{c}{2,562,921.5} \\
        MWU P-value & \multicolumn{2}{c}{0.012} & \multicolumn{2}{c}{$<10^{-3}$} \\
        KS Statistic & \multicolumn{2}{c}{0.0702} & \multicolumn{2}{c}{0.146} \\
        KS P-value & \multicolumn{2}{c}{$<10^{-3}$} & \multicolumn{2}{c}{$<10^{-3}$} \\
        \bottomrule
    \end{tabular}
\end{table}

These results demonstrate that the MEBeauty-trained model fails to satisfy both Distributional Parity and Error Parity criteria when applied to the SCUT-FBP5500 dataset, providing compelling evidence of ethnic bias in cross-dataset applications of beauty prediction models.

\subsubsection{FairFace Data Analysis}

\begin{figure}[t]
    \centering
    \includegraphics[width=\linewidth]{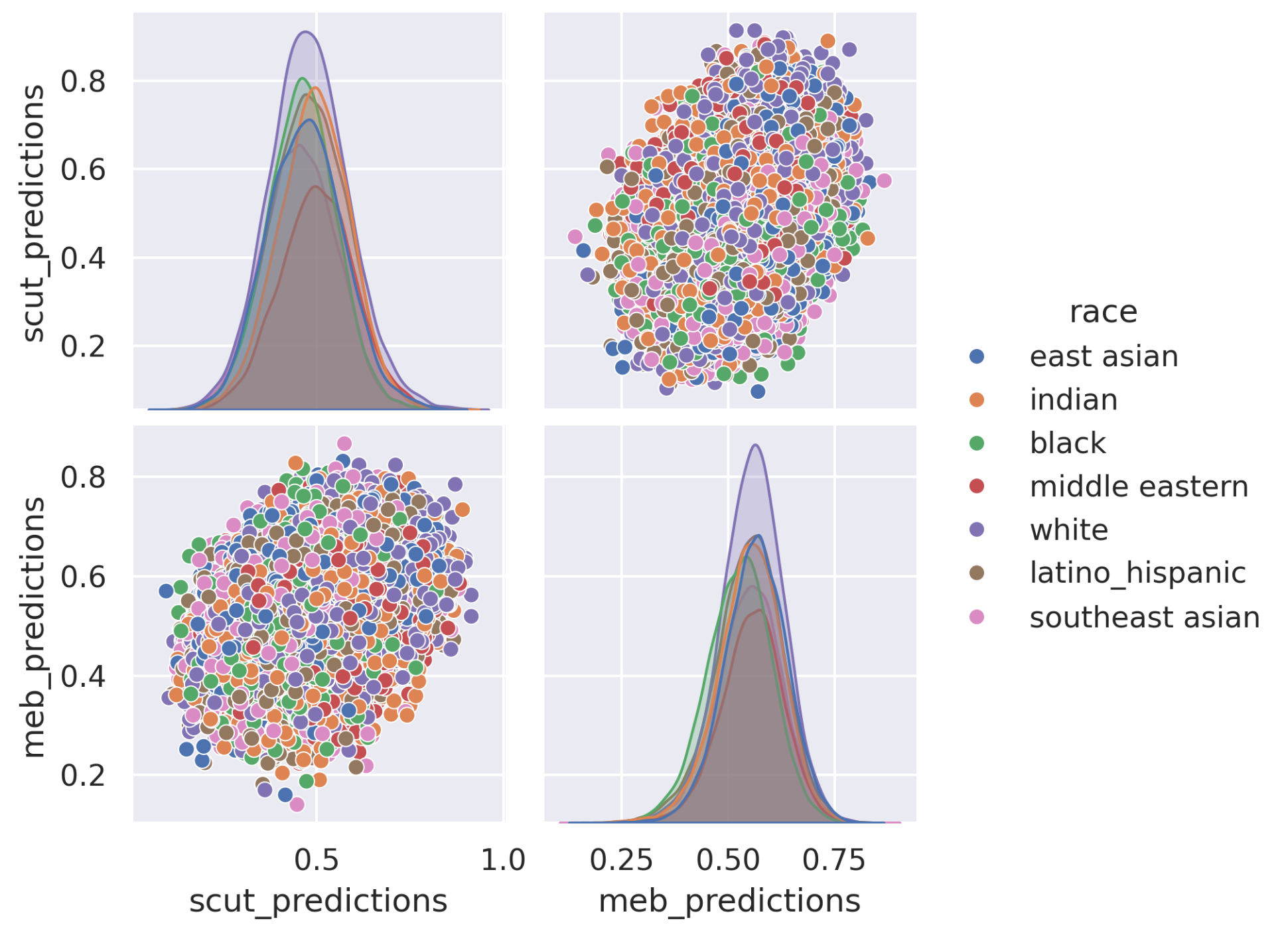}
    \caption{Pairplot of beauty score predictions from SCUT-trained and MEBeauty-trained models applied to the FairFace dataset, segmented by ethnic group. Differences in distribution shapes and centers reveal systematic prediction disparities across ethnicities.}
    \label{fig:fairface-pairplot}
\end{figure}

To assess bias generalization to a more diverse population, both trained models were evaluated on the demographically balanced FairFace dataset. Fig.~\ref{fig:fairface-pairplot} presents a pairplot visualization comparing prediction distributions across seven ethnic groups. Visual inspection reveals slight patterns of differential treatment across ethnicities by both models. Though distribution overlap is significant given the volume of samples in such a narrow prediction window, statistical tests reveal robust disparities in predictions across groups for both models.

Table~\ref{table:fairface-data} quantifies these disparities through descriptive statistics. Kruskal-Wallis tests strongly reject the null hypothesis of equal prediction distributions across ethnic groups for both the SCUT-trained model ($H = 1675.7$, $p < 10^{-3}$) and MEBeauty-trained model ($H = 1716.8$, $p < 10^{-3}$). These significant results confirm systematic differential treatment by race in both models.

\begin{table}[t]
\centering
\caption{Comparison of beauty score predictions from SCUT-trained and MEBeauty-trained models on the FairFace dataset across seven ethnic groups. Lower rows display Kruskal-Wallis test results assessing equality of prediction distributions.}
\label{table:fairface-data}
    \begin{tabular}{lcccc}
        \toprule
         & \multicolumn{2}{c}{SCUT Model} & \multicolumn{2}{c}{MEBeauty Model} \\
        \cmidrule(lr){2-3} \cmidrule(lr){4-5}
        Ethnic Group & Mean & Median & Mean & Median \\
        \midrule
        Black & 0.460 & 0.461 & 0.525 & 0.527 \\
        East Asian & 0.470 & 0.469 & 0.563 & 0.566 \\
        Indian & 0.493 & 0.494 & 0.550 & 0.555 \\
        Hispanic & 0.479 & 0.479 & 0.548 & 0.551 \\
        Middle Eastern & 0.500 & 0.500 & 0.555 & 0.559 \\
        Southeast Asian & 0.455 & 0.455 & 0.547 & 0.551 \\
        White & 0.479 & 0.477 & 0.556 & 0.559 \\
        \midrule
        KW Test Statistic & \multicolumn{2}{c}{1675.7} & \multicolumn{2}{c}{1716.8} \\
        KW Test P-value & \multicolumn{2}{c}{$<10^{-3}$} & \multicolumn{2}{c}{$<10^{-3}$} \\
        \bottomrule
    \end{tabular}
\end{table}

\begin{figure}[t]
    \centering
    \includegraphics[width=\linewidth]{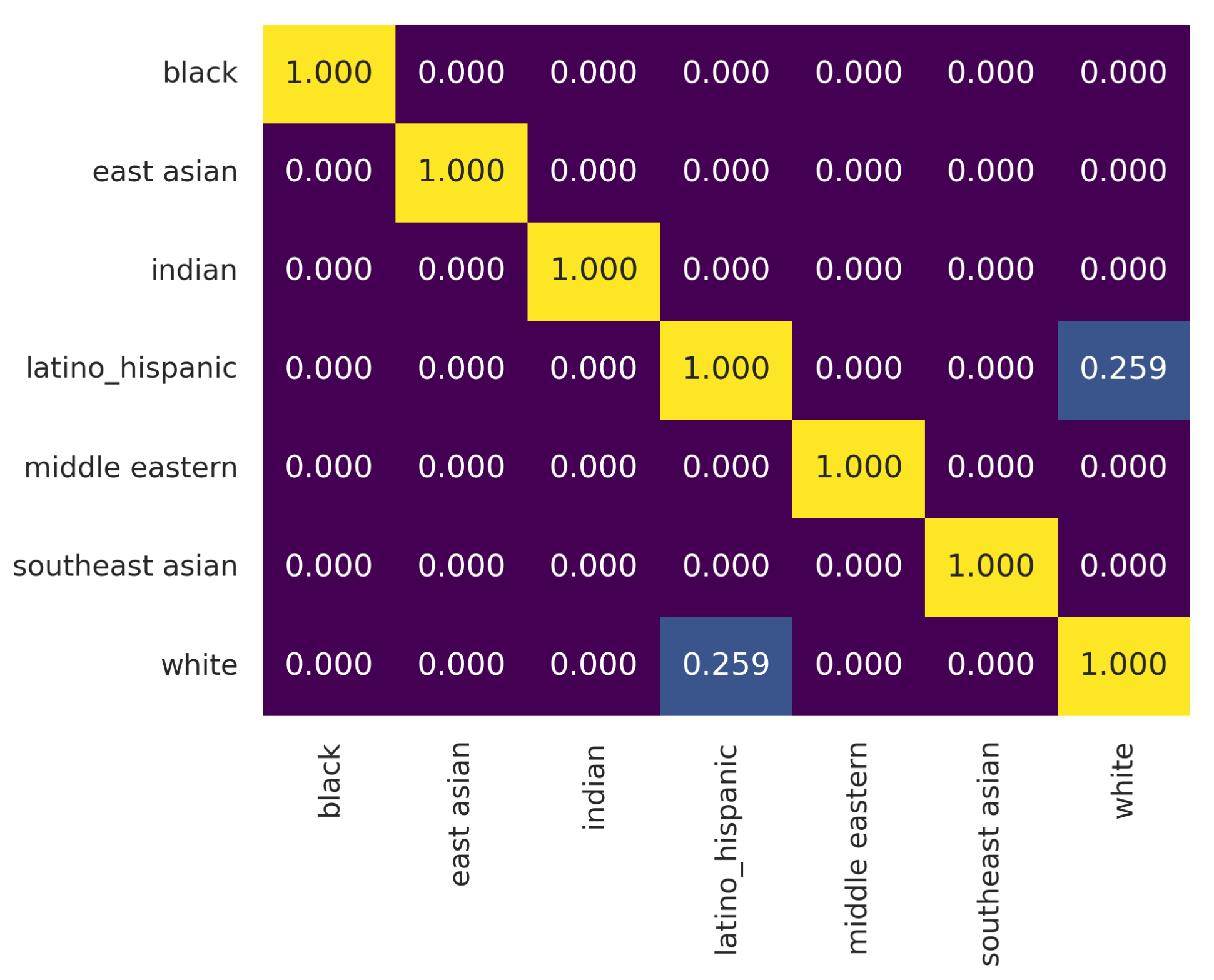}
    \caption{Heatmap of pairwise post hoc Dunn test results for SCUT-trained model predictions across ethnic groups on FairFace data. Each cell displays the adjusted p-value after Benjamini-Hochberg correction.}
    \label{fig:fairface-phd-scut-preds}
\end{figure}

\begin{figure}[t]
    \centering
    \includegraphics[width=\linewidth]{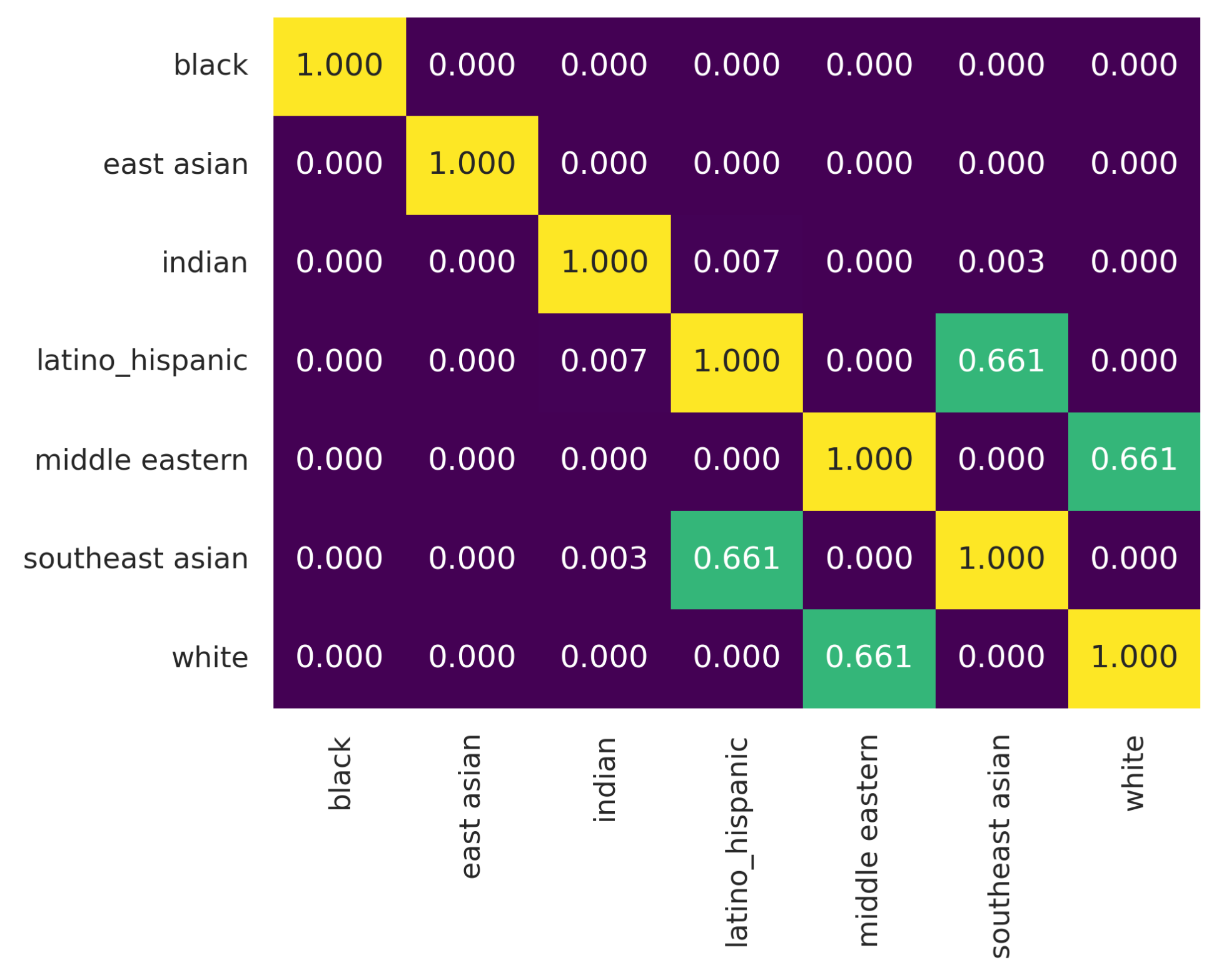}
    \caption{Heatmap of pairwise post hoc Dunn test results for MEBeauty-trained model predictions across ethnic groups on FairFace data. Each cell displays the adjusted p-value after Benjamini-Hochberg correction.}
    \label{fig:fairface-phd-meb-preds}
\end{figure}

Post Hoc Dunn tests with Benjamini-Hochberg corrections reveal the specific inter-group disparities. Fig.~\ref{fig:fairface-phd-scut-preds} shows that for the SCUT-trained model, only 1 out of 21 pairwise comparisons (4.8\%) satisfies Distributional Parity at $\alpha = 0.05$. Similarly, Fig.~\ref{fig:fairface-phd-meb-preds} demonstrates that the MEBeauty-trained model upholds Distributional Parity in only 2 of 21 comparisons (9.5\%). This analysis on a balanced dataset provides evidence that beauty prediction models systematically encode and reproduce ethnic biases.

\section{Discussion}

This study reveals ethnic bias in beauty prediction models, challenging an initial hypothesis that MEBeauty's broader ethnic representation would yield fairer models than SCUT-FBP5500. This finding aligns with \textcite{Chekanov2023HowAIAdoptsBias}'s framework for algorithmic bias propagation, where multiple systemic factors compound to produce discriminatory outcomes.

\subsection{Sources of Observed Bias}

- \textbf{Sampling Bias:} As visualized in Fig.~\ref{fig:meb-pairplot} and Fig.~\ref{fig:scut-pairplot}, both training datasets exhibit skewed label distributions across ethnic groups. SCUT-FBP5500's Asian/Caucasian dichotomy and MEBeauty's uneven score distributions suggest underlying sampling imbalances in original data collection.

- \textbf{Labeling Bias:} The consistent error disparities (Tables~\ref{table:meb-data}-\ref{table:scut-data}) point to systemic labeling inconsistencies. Even with multi-rater annotations, the lack of demographic parity and transparency among annotators likely introduced preference patterns into the training labels.

\subsection{Technical and Ethical Implications}

- \textbf{Model Generalization:} Both models exhibited exacerbated bias on the balanced FairFace dataset (Table~\ref{table:fairface-data}), suggesting that beauty prediction systems potentially amplify rather than mitigate existing societal biases when applied to diverse populations.

- \textbf{Fairness-Aware Production:} While the fairness criteria---requiring distributional parity across all ethnic comparisons---may seem stringent, this standard is necessary because even slight disparities can have outsized negative impacts on marginalized groups and should be the goal for responsible AI deployment.

\subsection{Paths Forward}

- \textbf{Algorithmic Mitigation:} Integration of fairness constraints during training, such as the frameworks proposed by \textcite{yik2024enforcing} and \textcite{yazdani2024fairbinn}, could help promote fairness along with model performance.

- \textbf{Data Curation:} Future datasets should enforce demographic balance through stratified sampling, implement annotator diversity quotas mirroring target populations, and include metrics to capture cultural relativity and representation

- \textbf{Validation Protocols:} Before model deployment addtional phases of model validation should be added to test for bias and provide transparent reporting of sampling, labeling, and disparity metrics across groups.

The road to equitable beauty AI remains challenging, but these results demonstrate that current approaches risk cementing harmful stereotypes rather than advancing inclusivity. The normalization of algorithmic beauty standards threatens to eclipse humanity's rich diversity of self-expression.

\section{Conclusion}

This study demonstrates that facial beauty prediction models have the potential to systematically encode ethnic biases, with both SCUT- and MEBeauty-trained models showing significant disparities across groups $(p < 0.001)$. The cross-dataset validation on FairFace revealed exacerbated bias in balanced populations, indicating amplification rather than mitigation of societal prejudices. These biases stem from compounded representation and annotation limitations in beauty datasets, where sampling imbalances and biased ratings propagate through model training.

Beauty AI systems require fairness-constrained architectures \parencite{yazdani2024fairbinn}, demographic equity validation protocols, and ethically curated datasets with diverse annotators. Unchecked deployment risks cementing algorithmic beauty standards that erase cultural diversity. Future work must prioritize decoupling aesthetic assessment from ethnic features while preserving model utility. This will be a critical step toward equitable AI in a socially impactful application.

\printbibliography

@inproceedings{Quer2024Analyzing,
title = {Analyzing bias and discrimination in an algorithmic hiring use case},
author = {David Quer and Anna Via and Marc Serra-Vidal and Laia Nadal and Didac Fortuny},
booktitle = {AIMMES 2024 Workshop on AI bias: Measurements, Mitigation, Explanation Strategies, co-located with EU Fairness Cluster Conference 2024},
year = {2024},
address = {Amsterdam, Netherlands},
}

@incollection{Bogdanova2024Diversity,
  author    = {Daryna Bogdanova and Anthi Vasiliki Haritos and Louisa Kieschnick},
  title     = {Assessing Diversity in AI-Generated Images for Beauty Campaigns},
  booktitle = {Special Issue Innovative Brand Management IV},
  editor    = {Carsten Baumgarth},
  series    = {Working Papers Innovative Brand Management},
  number    = {6},
  year      = {2024},
  pages     = {1--28},
  publisher = {Berlin School of Economics and Law (HWR Berlin)},
  address   = {Berlin, Germany},
  issn      = {2702-1491}
}

@inproceedings{Gengler2024Sexism,
  author    = {Gengler, Eva Johanna},
  title     = {Sexism, Racism, and Classism: Social Biases in Text-to-Image Generative AI in the Context of Power, Success, and Beauty},
  booktitle = {Wirtschaftsinformatik 2024 Proceedings},
  year      = {2024},
  pages     = {48},
  url       = {https://aisel.aisnet.org/wi2024/48}
}

@article{Yang2025Racial,
  author    = {Yiran Yang},
  title     = {Racial bias in AI-generated images},
  journal   = {AI \& Society},
  year      = {2025},
  doi       = {10.1007/s00146-025-02282-1},
  note      = {Open Access, Creative Commons Attribution 4.0 International License},
  publisher = {Springer},
  address   = {London},
  url       = {https://doi.org/10.1007/s00146-025-02282-1}
}

@article{Feldman2021EndToEnd,
  title     = {End-To-End Bias Mitigation: Removing Gender Bias in Deep Learning},
  author    = {Tal Feldman and Ashley Peake},
  journal   = {arXiv preprint arXiv:2104.02532},
  year      = {2021},
  url       = {https://arxiv.org/abs/2104.02532}
}

@mastersthesis{Ilyas2024AIBeauty,
  author    = {Qudsia Ilyas},
  title     = {AI in Beauty Content: A Theoretical Study of Ethical and Psychological Considerations Surrounding AI-Generated Beauty Content},
  school    = {Vaasa University of Applied Sciences},
  type      = {Bachelor's Thesis},
  year      = {2024},
  address   = {Vaasa, Finland},
  supervisor= {Jaana Ylimartimo-Nyback and Johanna Nykamb},
  note      = {Degree Program of Beauty and Cosmetics - Beauty Care (UAS)},
  url       = {https://www.theseus.fi/handle/10024/872723}
}

@unpublished{Zhou2024AIBeauty,
  author    = {Belle Zhou},
  title     = {AI and Human Beauty Standards},
  year      = {2024},
  note      = {Manuscript},
  url       = {https://philarchive.org/rec/ZHOAAH}
}

@article{He2016Deep,
  title     = {Deep Residual Learning for Image Recognition},
  author    = {Kaiming He and Xiangyu Zhang and Shaoqing Ren and Jian Sun},
  journal   = {arXiv preprint arXiv:1512.03385},
  year      = {2015},
  url       = {https://arxiv.org/abs/1512.03385}
}

@article{liang2017SCUT,
  title     = {SCUT-FBP5500: A Diverse Benchmark Dataset for Multi-Paradigm Facial Beauty Prediction},
  author    = {Liang, Lingyu and Lin, Luojun and Jin, Lianwen and Xie, Duorui and Li, Mengru},
  journal    = {ICPR},
  year      = {2018}
}

@article{lebedeva2021mebeauty,
  title={MEBeauty: a multi-ethnic facial beauty dataset in-the-wild},
  author={Lebedeva, Irina and Guo, Yi and Ying, Fangli},
  journal={Neural Computing and Applications},
  pages={1--15},
  year={2021},
  publisher={Springer}
}

@inproceedings{karkkainenfairface,
    title={FairFace: Face Attribute Dataset for Balanced Race, Gender, and Age for Bias Measurement and Mitigation},
    author={Karkkainen, Kimmo and Joo, Jungseock},
    booktitle={Proceedings of the IEEE/CVF Winter Conference on Applications of Computer Vision},
    year={2021},
    pages={1548--1558}
}

@article{Thomee_2016,
   title={YFCC100M: the new data in multimedia research},
   volume={59},
   ISSN={1557-7317},
   url={http://dx.doi.org/10.1145/2812802},
   DOI={10.1145/2812802},
   number={2},
   journal={Communications of the ACM},
   publisher={Association for Computing Machinery (ACM)},
   author={Thomee, Bart and Shamma, David A. and Friedland, Gerald and Elizalde, Benjamin and Ni, Karl and Poland, Douglas and Borth, Damian and Li, Li-Jia},
   year={2016},
   month=jan, pages={64–73} 
}

@INPROCEEDINGS{5206848,
  author={Deng, Jia and Dong, Wei and Socher, Richard and Li, Li-Jia and Kai Li and Li Fei-Fei},
  booktitle={2009 IEEE Conference on Computer Vision and Pattern Recognition}, 
  title={ImageNet: A large-scale hierarchical image database}, 
  year={2009},
  volume={},
  number={},
  pages={248-255},
  doi={10.1109/CVPR.2009.5206848}}

@article{Chekanov2023HowAIAdoptsBias,
  title     = {How artificial intelligence adopts human biases: the case of cosmetic skincare industry},
  author    = {Konstantin Chekanov and Anastasia Georgievskaya and Timur Tlyachev and Daniil Danko and Hugo Corstjens},
  journal   = {npj Digital Medicine},
  year      = {2023},
  volume    = {6},
  number    = {1},
  pages     = {194},
  url       = {https://link.springer.com/article/10.1007/s43681-023-00378-2}
}

@article{yazdani2024fairbinn,
  title={Fair Bilevel Neural Network (FairBiNN): On Balancing Fairness and Accuracy via Stackelberg Equilibrium},
  author={Yazdani-Jahromi, Mehdi and Yalabadi, Ali Khodabandeh and Rajabi, Amirarsalan and Tayebi, Aida and Garibay, Ivan and Garibay, Ozlem},
  journal={arXiv preprint arXiv:2410.16432},
  year={2024},
  url={https://arxiv.org/abs/2410.16432}
}

@article{Zhang_2016,
   title={Joint Face Detection and Alignment Using Multitask Cascaded Convolutional Networks},
   volume={23},
   ISSN={1558-2361},
   url={http://dx.doi.org/10.1109/LSP.2016.2603342},
   DOI={10.1109/lsp.2016.2603342},
   number={10},
   journal={IEEE Signal Processing Letters},
   publisher={Institute of Electrical and Electronics Engineers (IEEE)},
   author={Zhang, Kaipeng and Zhang, Zhanpeng and Li, Zhifeng and Qiao, Yu},
   year={2016},
   month=oct, pages={1499–1503} }

@article{yik2024enforcing,
  title = {Enforcing Equity in Neural Climate Emulators},
  author = {Yik, William and Silva, Sam J.},
  journal = {arXiv preprint arXiv:2406.19636},
  year = {2024},
  url = {https://arxiv.org/abs/2406.19636}
}

@misc{xie2017aggregatedresidualtransformationsdeep,
      title={Aggregated Residual Transformations for Deep Neural Networks}, 
      author={Saining Xie and Ross Girshick and Piotr Dollár and Zhuowen Tu and Kaiming He},
      year={2017},
      eprint={1611.05431},
      archivePrefix={arXiv},
      primaryClass={cs.CV},
      url={https://arxiv.org/abs/1611.05431}, 
}
\end{document}